\documentclass{article}
\usepackage{ijcai19}

\usepackage{times}
\usepackage{soul}
\usepackage{url}
\usepackage[hidelinks]{hyperref}
\usepackage[utf8]{inputenc}
\usepackage[small]{caption}
\usepackage{graphicx}
\usepackage{amsmath}
\usepackage{booktabs}
\usepackage{color}
\usepackage{multicol}
\usepackage{amsmath,epsfig,bm, amssymb,subfigure,graphicx,algorithm,algorithmic,color, bbm}

\newcommand{\argmin}{\mathop{\mathrm{argmin\,}}}
\newcommand{\argmax}{\mathop{\mathrm{argmax\,}}}

\newcommand{\mathbbR}{\mathbb{R}}

\newcommand{\boldx}{{\boldsymbol{x}}}
\newcommand{\boldy}{{\boldsymbol{y}}}

\newcommand{\boldmu}{{\boldsymbol{\mu}}}

\newcommand{\calD}{{\mathcal{D}}}

\newcommand{\calH}{{\mathcal{H}}}

\usepackage{amsfonts}
\usepackage{braket}
\usepackage{bm}

\title{Topological Bayesian Optimization with Persistence Diagrams }

\author{
Tatsuya Shiraishi$^1$\and
Tam Le$^2$\and
Hisashi Kashima$^{1,2}$\and
Makoto Yamada$^{1,2}$\footnote{Contact Author}\\
\affiliations
$^1$Kyoto University, $^2$RIKEN AIP\\
\emails
\texttt{shiraishi.t@ml.ist.i.kyoto-u.ac.jp}, \texttt{tam.le@riken.jp}, \texttt{\{kashima,myamada\}@i.kyoto-u.ac.jp}
}

\begin{document}

\maketitle

\begin{abstract}
Finding an optimal parameter of a black-box function is important for searching stable material structures and finding optimal neural network structures, and Bayesian optimization algorithms are widely used for the purpose. 
However, most of existing Bayesian optimization algorithms can only handle vector data and cannot handle complex structured data. 
In this paper, we propose the topological Bayesian optimization, which can efficiently find an optimal solution from structured data using \emph{topological information}.
More specifically, in order to apply Bayesian optimization to structured data, we extract useful topological information from a structure and measure the proper similarity between structures.
To this end, we utilize persistent homology, which is a topological data analysis method that was recently applied in machine learning.
Moreover, we propose the Bayesian optimization algorithm that can handle multiple types of topological information by using a linear combination of kernels for persistence diagrams. 
Through experiments, we show that topological information extracted by persistent homology contributes to a more efficient search for optimal structures compared to the random search baseline and the graph Bayesian optimization algorithm.
\end{abstract}

\section{Introduction}
In recent years, many studies have been actively conducted on the analysis of data with complex structures like graph structures. 
Graph structure optimization involves searching for graph structures with optimal properties, and it is one of the fundamental tasks in graph structured data analysis.
Examples of graph structure optimization include searching for stable lowest-energy crystal structures \cite{calypso} and searching for road networks with optimal traffic volume \cite{utndp}. 
Another example of graph structure optimization would be neural network architecture search \cite{nasbot}, which is an important task in deep learning architecture research.
Thus, learning from complex structure including graphs is very important in various research fields.

\par
The objective function of structure optimization (e.g., energy of a crystal structure and traffic volume of a road network) is an expensive-to-evaluate function, which needs to be measured by performing a long time experiment or a large scale investigation, and is a black-box function, which cannot be written explicitly. 
Therefore, an optimization method that can optimize even an unknown objective function with fewer evaluations of the function is desirable. 
Bayesian optimization is one of methods that satisfies this condition. 
However, studies on Bayesian optimization often assume vector data as the input, and few studies have focused on structured data.
In standard Bayesian optimization methods, we tend to use the Gaussian kernel function, which expresses the similarity between input vectors.
Thus, to handle structured data (e.g., graphs) by Bayesian optimization, we need to design a similarity that properly captures the structure.
For example, a method using graph kernels was proposed for handling arbitrary graph structures by Bayesian optimization \cite{ramachandram2018bayesian,cui2018graph}, and this method outperforms vector based Bayesian optimization in tasks such as identifying the most active node in a social network and searching for optimal transportation networks.

\par
Recently, the topological data analysis (TDA) has received considerable attention in machine learning as a technique for extracting topological features from complex structured data. 
Persistent homology is a TDA method that is actively studied for application to statistical machine learning.
This method extracts topological features from a point cloud on a metric space, and the result is represented by a point cloud on $\mathbb{R}^2$ called a persistence diagram (PD).
As one of the applications of persistent homology to machine learning, several kernels for PD have been proposed, and the effectiveness has been demonstrated by classification tasks using the support vector machines (SVM) and change point detection \cite{kusano2018kernel,le2018persistence}.
However, to the best of our knowledge, there is no Bayesian optimization method that utilizes topological data analysis.

\par
In this paper, we propose the topological Bayesian optimization, which is a Bayesian optimization algorithm using features extracted by persistent homology.
More specifically, we first introduce the persistence weighted Gaussian kernel (PWGK) \cite{kusano2018kernel} and the persistence Fisher kernel (PFK) \cite{le2018persistence} for Gaussian processes, and derive a Bayesian optimization algorithm for topological information.
Since the current persistence homology based approach considers only one type of topological information, it may not be able to capture various types of topological information. 
Therefore, we further propose a multiple kernel learning based algorithm and apply it to Bayesian optimization problems.
Through experiments using synthetic and two real datasets, we show that our method can search for the optimal structure more efficiently compared to the random search baseline and the state-of-the-art Bayesian optimization for graphs \cite{ramachandram2018bayesian,cui2018graph}.

\vspace{.1in}
\noindent {\bf Contributions:} The contributions of this paper are summarized as follows:
\begin{itemize}
    \item We propose a the Bayesian optimization algorithm utilizing topological data analysis.
    \item We further propose a multiple kernel learning based algorithm to use various types of topological information.
    \item Through experiments, we show that our method can search for the optimal structure more efficiently compared to the random search baseline and the graph Bayesian optimization algorithms.
\end{itemize}


\section{Background}
In this section, we briefly review the traditional Bayesian optimization algorithms based on Gaussian process and the topological data analysis (TDA).

\subsection{Bayesian optimization}
Bayesian optimization is an effective optimization method for expensive-to-evaluate objective functions \cite{bo}.
Let us denote the input vector $\boldx \in \mathbbR^d$ and a black box function $f: \mathbbR^d \rightarrow \mathbbR$. 
Bayesian optimization tries to find the optimal data point of the following optimization problem:
\begin{align*}
    \boldx^\ast = \argmin_{\boldx \in \mathbbR^d}~f(\boldx).
\end{align*}
Since Bayesian optimization does not need derivatives for finding the optimal data point, it is particularly effective when optimizing black-box objective functions.
Bayesian optimization is an iterative method, and each step consists of two steps: (i) calculation of a predictive distribution of an objective function value by a Gaussian process and (ii) selection of the next search point based on an acquisition function.

\vspace{.1in}
\noindent {\bf Gaussian process:} Gaussian process is a generalization of Gaussian probability distribution \cite{gp}. 
More specifically, Gaussian process describes the functions of random variables, while Gaussian probability distribution describes random scalars or vectors.
In Bayesian optimization, the objective function $f:\mathbbR^d\to\mathbb{R}$ is modeled by a Gaussian process, which enables easy calculation of predictive distributions.
Now, let $\mathcal{X}=\set{(\bm{x}_1,y_1),\cdots,(\bm{x}_t,y_t)}$ be pairs of the input and the corresponding output of the objective function observed up to a certain step.
Here, the true value $f(\bm{x}_i)$ is not necessarily observed as $y_i$, but an independent additive Gaussian noise $\epsilon_i\sim\mathcal{N}(0,\sigma^2)$ is included:
\begin{align*}
    y_i = f(\bm{x}_i)+\epsilon_i.
\end{align*}
According to the definition of Gaussian process, the joint probability distribution of $f(\bm{x}_1),\cdots,f(\bm{x}_t)$ is 
\begin{align}
    (f(\bm{x}_1),\cdots,f(\bm{x}_t))^T\sim\mathcal{N}(\bm{0},\bm{K}), \label{joint}
\end{align}
where $\bm{0}=(0,\cdots,0)^T$, $\cdot^T$ denotes the transpose operator, and each element of $\bm{K}\in\mathbb{R}^{t\times t}$ is expressed by $K_{ij}=k(\bm{x}_i,\bm{x}_j)$ using the kernel function $k(\cdot,\cdot)$.
Then, the predictive distribution of the function value $f(\bm{x}_{t+1})$ at the point $\bm{x}_{t+1}$, which is not included in the data, can be calculated.
Since the joint probability distribution of $f(\bm{x}_1),\cdots,f(\bm{x}_t),f(\bm{x}_{t+1})$ is also expressed similar to the expression (i.e., Eq.~\eqref{joint}) and the additive noise is included in the observations, the predictive distribution of $f(\bm{x}_{t+1})$ is also a Gaussian distribution whose mean $\mu(\bm{x}_{t+1})$ and covariance $\sigma^2(\bm{x}_{t+1})$ are as follows:
\begin{align*}
    \mu(\bm{x}_{t+1}) &= \bm{k}(\bm{K}+\sigma^2\bm{I})^{-1}\bm{y}, \\
	\sigma^2(\bm{x}_{t+1}) &= k(\bm{x}_{t+1},\bm{x}_{t+1})-\bm{k}(\bm{K}+\sigma^2\bm{I})^{-1}\bm{k}^T,
\end{align*}
where $\bm{k}=(k(\bm{x}_{t+1},\bm{x}_1),\cdots,k(\bm{x}_{t+1},\bm{x}_{t}))$ and $\bm{y}=(y_1,\cdots,y_t)^T$. (See \cite{gp} for the detailed derivation).

\vspace{.1in}
\noindent {\bf Acquisition function:} The acquisition function ${\rm acq}(\bm{x})$ expresses the degree to which we should evaluate the input point $\bm{x}$ based on the predictive distribution calculated utilizing a Gaussian process.
In Bayesian optimization, the point that maximizes the acquisition function is selected as the next evaluation point:
\begin{align*}
    \bm{x}_{t+1} = \argmax_{\bm{x}\in\mathbbR^d}~{\rm acq}(\bm{x}).
\end{align*}
There are many acquisition functions including probability of improvement (PI) \cite{pi}, expected improvement (EI) \cite{ei}, and lower confidence bound (LCB) \cite{lcb}.
The balance between exploitation and exploration is important for acquisition functions.
Exploitation involves evaluation of points in the surroundings of the point observed with the best objective function value, while exploration involves evaluation of points with high uncertainty.
EI, which we use in the experiments, is the expected value of the difference between the best observation value $y_{best}$ obtained up to a certain step and the predicted objective function value $f(\bm{x})$.
\begin{align*}
    {\rm acq}_{EI}(\bm{x}) &= \mathbb{E}[\max\{0, y_{best}-f(\bm{x})\}] \nonumber\\
    & = \begin{cases}
        \sigma(\bm{x})(Z\Phi(Z)+\phi(Z)) & \sigma(\bm{x})\neq 0 \\
        0 & \sigma(\bm{x})=0
        \end{cases},
\end{align*}
where $Z=\frac{y_{best}-\mu(\bm{x})}{\sigma(\bm{x})}$, and $\Phi$ and $\phi$ are the cumulative density function and probability density function of a standard normal distribution, respectively.

\subsection{TDA based on persistent homology}
\begin{figure}[htbp]
	\centering
	\includegraphics[clip, width=0.48\textwidth, trim=4cm 12cm 3cm 12cm]{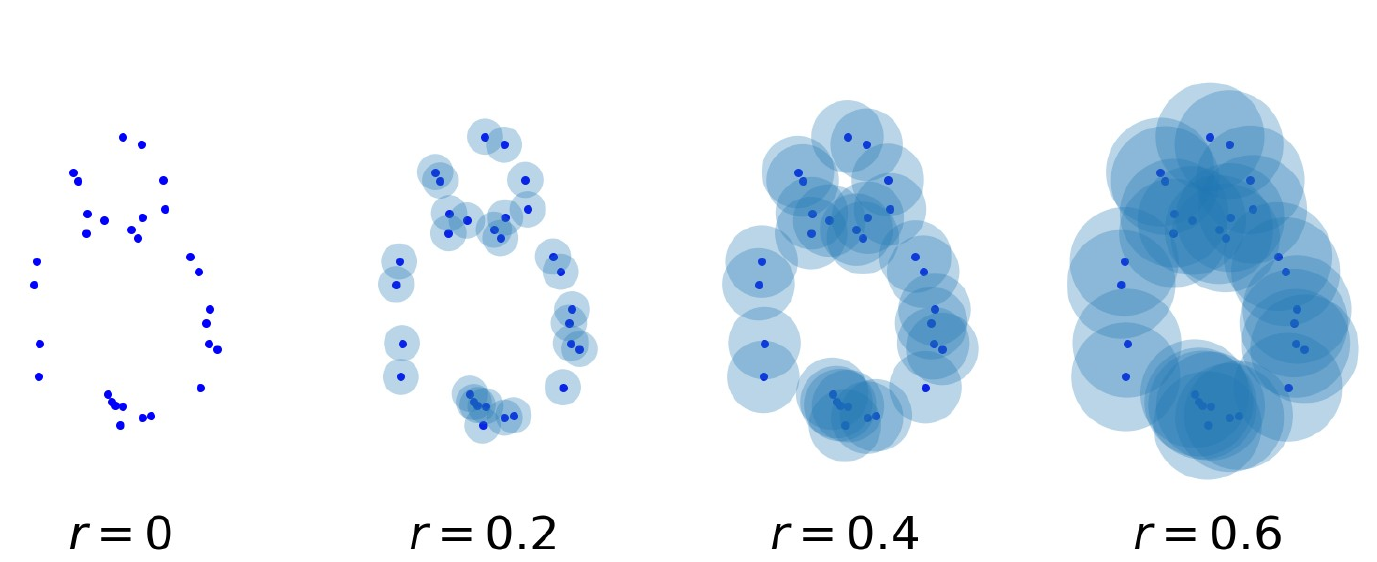}
	\caption{Examples of $S_r$.}
	\label{filtration}
	\vspace{-.2in}
\end{figure}

In TDA, we focus on the shapes of a complex data represented by a point cloud or a graph from the viewpoint of topology.
Here, we give an intuitive explanation of one of the TDA methods, namely persistent homology \cite{ph}.
In order to analyze a point cloud $\set{\boldx_1,\cdots,\boldx_N}$ on a metric space $(M,c)$ by persistent homology, we consider the union of balls centered on each point with radius $r$:
\begin{align*}
    S_r = \bigcup_{i=1}^N\set{\boldx\in M|c(\boldx,\boldx_i)\le r}.
\end{align*}
Figure \ref{filtration} shows examples of $S_r$.
We can observe that topological structures like connected components and rings appear and disappear.
In persistent homology, we focus on when each topological structure appears and how long it persists.

\par
\begin{figure}[htbp]
	\centering
	\includegraphics[clip, width=0.48\textwidth, trim=2cm 10cm 2cm 10cm]{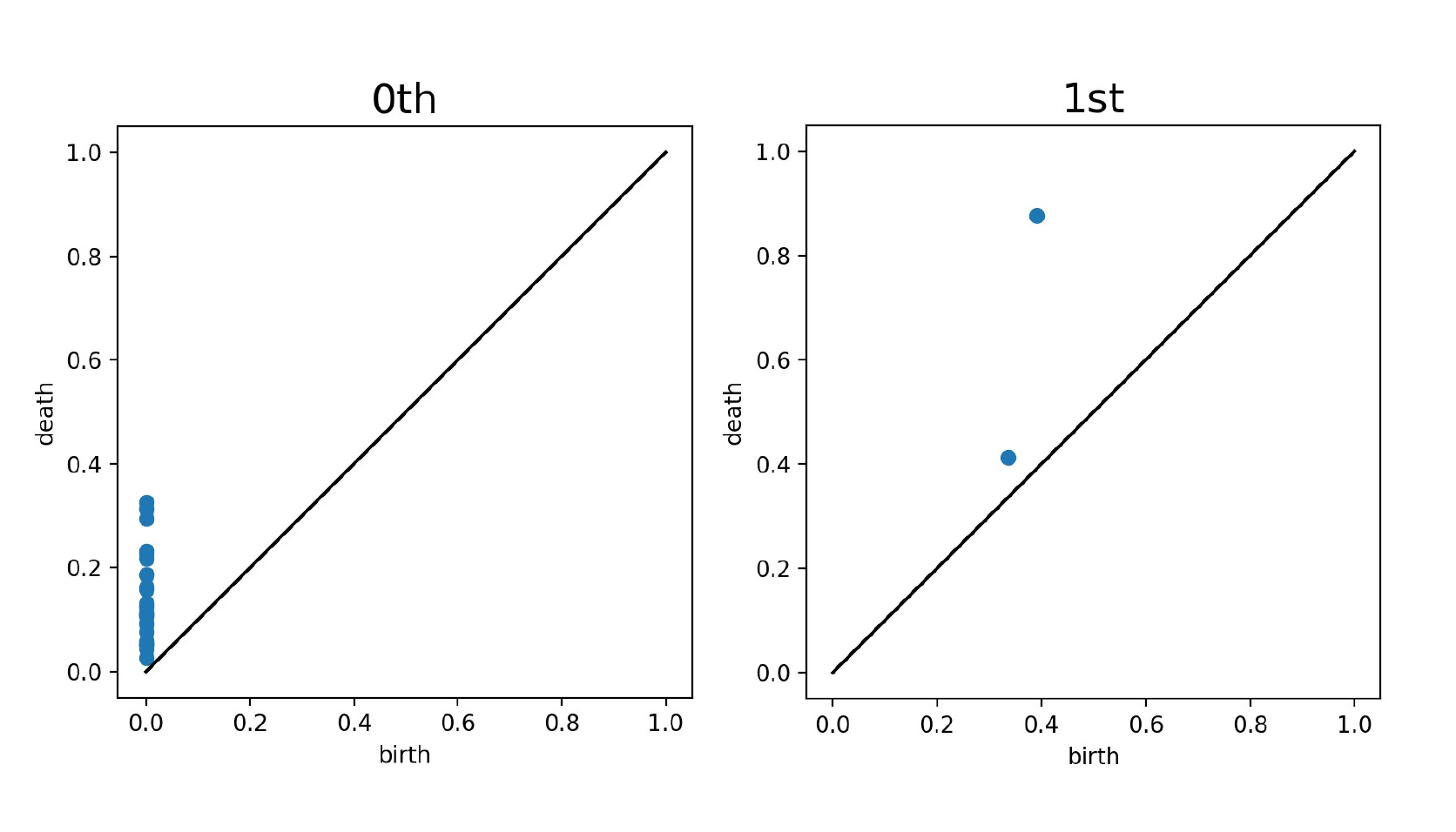}
	\caption{0th and 1st PDs for the point cloud of Figure \ref{filtration}.}
	\label{pd}
	\vspace{-.2in}
\end{figure}

The topological features extracted by persistent homology can be expressed as a point cloud on $\mathbb{R}^2$ called a persistence diagram (PD).
A point $(b,d)$ on a PD shows the corresponding topological structure that appears at radius $b$ and disappears at radius $d$.
Since $b<d$, all the points on a PD are distributed above the diagonal.
We can consider multiple PDs for the same point cloud depending on the structure of interest.
It is called the 0th PD when we focus on the connected components, the 1st PD when we focus on the rings and so on.
Figure \ref{pd} shows the 0th PD and the 1st PD for the point cloud of Figure \ref{filtration}.
Two points corresponding to the large ring and the small ring in the point cloud can be seen in the 1st PD.
The smaller ring corresponds to the point closer to the diagonal, while the larger ring corresponds to the point farther from the diagonal.
Thus, the points distributed near the diagonal may represent noisy structures that disappear quickly, while the points distributed far from the diagonal may represent more important structures.

\section{Proposed method: Topological Bayesian optimization}
In order to handle structured data by Bayesian optimization, it is necessary to design a similarity that captures the topological features of a structure.
Although TDA has attracted considerable attention as techniques that can extract such features from complex data, there has been no Bayesian optimization method utilizing TDA to design the similarity. 
Therefore, in this paper, we propose Bayesian optimization utilizing features extracted by persistent homology. 

\par
Moreover, most studies on the applications of persistent homology to machine learning, especially studies on kernels for PDs, consider one type of PD extracted from one data to calculate the kernel. 
However, it is possible to extract multiple types of PD from one data by using persistent homology.
We further propose methods to handle multiple topological features extracted by persistent homology by constructing a kernel using kernels calculated from each type of PD. 

In this section, we first formulate the \emph{topological Bayesian optimization} problem using persistence diagrams. 
Then, we propose the kernel based Bayesian optimization algorithms. 

\subsection{Problem formulation}
Let us denote an input persistence diagram by $D_i$ and the set of persistence diagrams by $\mathcal{D}=\set{D_i}_{i\in I}$, where $I$ is the set of \emph{oracle} indices that we cannot observe in the beginning.
In addition, we assume that evaluating a persistence diagram $D_i$ is expensive. Since TDA is highly used in material science, this assumption is rather reasonable.  

\par
In this paper, we consider searching for the point that minimizes the objective function from $\calD$:
\begin{align}
\label{tda_formulation}
    D^\ast = \argmin_{D \in \calD}~f(D),
\end{align}
where $f(\cdot)$ is a black box function. 
This problem can be solved easily if we can examine all possible cases.
However, since the objective function is expensive to evaluate, we need to find the optimal data point with a small number of evaluations. 
Note that we assume that the objective function value can be observed only in a state of including the independent additive Gaussian noise $\epsilon_i\sim\mathcal{N}(0,\sigma^2)$.
The final goal of this paper is to develop a Bayesian optimization algorithm to solve Eq.~\eqref{tda_formulation}.




\par
We first introduce kernels for PDs in Section \ref{pd_kernels}, and then explain methods for constructing a kernel from multiple kernels in Section \ref{mkl}.

\subsection{Kernels for persistence diagrams\label{pd_kernels}}

\vspace{.1in}
\noindent {\bf Persistence weighted Gaussian kernel:} Persistence weighted Gaussian kernel (PWGK) considers a PD as a weighted measure.
It first vectorizes the measure on an RKHS by kernel mean embedding, and then uses conventional vectorial kernels (e.g., linear kernel and Gaussian kernel) on the RKHS \cite{kusano2018kernel}.
More specifically, it considers the following weighted measure for a persistent diagram $D$:
\begin{equation}
	\mu_D = \sum_{\boldx\in D}w(\boldx)\delta_\boldx,
\end{equation}
where $\delta_\boldx$ is a Dirac measure, which takes 1 for $\boldx$ and 0 for other points. 
Additionally Dirac measures are weighted by the weight function $w(\boldx): \mathbbR^2\to\mathbbR$ based on the idea that the points close to the diagonal in the PD may represent noisy features, while the points far from the diagonal may represent relatively important features.
Let $E(\mu_D)$ be the vector representation of $\mu_D$ embedded by kernel mean embedding into the RKHS $\mathcal{H}$. 
Then, the inner product (linear kernel) of persistence diagrams $D_i,D_j$ on the RKHS is as follows:
\begin{align*}
	k_L(D_i,D_j)
	& = \langle E(\mu_{D_i}),E(\mu_{D_j})\rangle_\mathcal{H} \\
	& = \sum_{\boldx\in D_i}\sum_{\boldy\in D_j}w(\boldx)w(\boldy)\exp\left(-\frac{\|\boldx-\boldy\|^2}{2\nu^2}\right),
\end{align*}
where $\nu > 0$ is the kernel bandwidth. 
In addition, the Gaussian kernel on the RKHS is as follows:
\begin{align*}
	k_G(D_i,D_j) = \exp\left(-\frac{\|E(\mu_{D_i})-E(\mu_{D_j})\|_\mathcal{H}^2}{2\tau^2}\right).
\end{align*}
We will refer to them as PWGK-Linear and PWGK-Gaussian, respectively. 
Here, $\tau>0$ and 
\begin{eqnarray*}
	&& \|E(\mu_{D_i})-E(\mu_{D_j})\|_\mathcal{H}^2 \\
	&& = \sum_{\boldx\in D_i}\sum_{\boldy\in D_i}w(\boldx)w(\boldy)\exp\left(-\frac{\|\boldx-\boldy\|^2}{2\nu^2}\right) \nonumber\\
	&& \phantom{=}+\sum_{\boldx\in D_j}\sum_{\boldy\in D_j}w(\boldx)w(\boldy)\exp\left(-\frac{\|\boldx-\boldy\|^2}{2\nu^2}\right) \nonumber\\
	&& \phantom{=}-2\sum_{\boldx\in D_i}\sum_{\boldy\in D_j}w(\boldx)w(\boldy)\exp\left(-\frac{\|\boldx-\boldy\|^2}{2\nu^2}\right).
\end{eqnarray*}
Note that PWGK can be efficiently computed by using random Fourier features \cite{rff}.

\vspace{.1in}
\noindent {\bf Persistence Fisher kernel:} Persistence Fisher kernel (PFK) \cite{le2018persistence} considers a PD as the sum of normal distributions and measures the similarity between the distributions by using the Fisher information metric.
Let $D_{i\Delta}$ and $D_{j\Delta}$ be the point sets obtained by projecting persistence diagrams $D_i$ and $D_j$ on the diagonal, respectively.
PFK compares $D_i'=D_i\cup D_{j\Delta}$ and $D_j'=D_j\cup D_{i\Delta}$ instead of comparing $D_i$ and $D_j$. 
It makes the sizes of each point cloud equal, which makes it easy to apply various similarities.
Then, it considers the following summation of normal distributions for $D_i'$:
\begin{equation*}
	\rho_{D_i'} = \frac{1}{Z}\sum_{\boldmu\in D_i'}\mathcal{N}(\boldmu,\nu\bm{I}),
\end{equation*}
where $Z=\int\sum_{\boldmu\in D_i'}\mathcal{N}(\boldx;\boldmu,\nu\bm{I})d\boldx$ is the normalization constant. 
The Fisher information metric of the probability distributions $\rho(D_i')$ and $\rho(D_j')$ is as follows:
\begin{equation*}
	d_{FIM}(D_i, D_j) = \arccos\left(\int\sqrt{\rho_{D_i'}(\boldx)\rho_{D_j'}(\boldx)}d\boldx\right).
\end{equation*}
The integral appearing in $Z$ and $d_{FIM}$ is calculated using the function value at $\Theta=D_i\cup D_{j\Delta}\cup D_j\cup D_{i\Delta}$. 
Finally, PFK is expressed as follows using the Fisher information metric:
\begin{equation*}
	k_{PF}(D_i, D_j) = \exp(-td_{FIM}(D_i,D_j)), 
\end{equation*}
where $t>0$ is the tuning parameter. 
Approximation of PFK using fast Gauss transform \cite{fgt} is also proposed.

\subsection{Multiple kernel learning\label{mkl}}
In order to handle multiple topological features, we construct an additive kernel calculated from each feature.
In particular, we consider a linear combination of $k$ Gram matrices $\bm{K}_1,\cdots,\bm{K}_k$:
\begin{equation}
	\bm{K} = \alpha_1\bm{K}_1+\cdots +\alpha_k\bm{K}_k,
	\label{combination}
\end{equation}
where $\alpha_i\ge 0$ for all $i$. 
This construction makes it possible to maintain the positive definiteness of each kernel.
We consider two methods to learn the coefficient parameter $\bm{\alpha}=(\alpha_1,\cdots,\alpha_k)^T$.

\vspace{.1in}
\noindent {\bf Kernel target alignment:} 
A method of maximizing a value called alignment was proposed to learn $\bm{\alpha}$ \cite{cortes2012algorithms}. 
It first considers the centered Gram matrix $\bm{K}_c$ for the Gram matrix $\bm{K}$:
\begin{equation*}
	(K_c)_{ij} = K_{ij}-\mathbb{E}_i[K_{ij}]-\mathbb{E}_j[K_{ij}]+\mathbb{E}_{i,j}[K_{ij}].
\end{equation*}
Then, the alignment of the two Gram matrices $\bm{K},\bm{K}'$ is defined as follows:
\begin{equation*}
	\kappa(\bm{K},\bm{K}') = \frac{\langle\bm{K}_c,\bm{K}'_c\rangle_F}{\|\bm{K}_c\|_F\|\bm{K}'_c\|_F},
\end{equation*}
where $\langle\cdot,\cdot\rangle_F$ is the Frobenius inner product and $\|\cdot\|_F$ is the Frobenius norm.
In the alignment-based method \cite{cortes2012algorithms}, we maximize the alignment of $\bm{K}=\sum_i\alpha_i\bm{K}_i$ and $\bm{Y}=\bm{y}\bm{y}^T$.
Maximization of the alignment results in the following quadratic programming problem:
\begin{equation*}
	\min_{\bm{v}\ge\bm{0}}\bm{v}^T\bm{M}\bm{v}-2\bm{v}^T\bm{a},
\end{equation*}
where 
\begin{eqnarray*}
	 M_{ij}\!=\!\langle\bm{K}_{ic},\bm{K}_{jc}\rangle_F,~~ \bm{a}\!=\!(\langle\bm{K}_{1c},\bm{Y}\rangle_F,\cdots,\langle\bm{K}_{kc},\bm{Y}\rangle_F)^T.
\end{eqnarray*}
Let $\bm{v}^*$ be the solution of this problem.
Then, the coefficients are calculated by $\bm{\alpha}=\bm{v}^*/\|\bm{v}^*\|$. 
Since $\bm{y}$ is updated at each step in Bayesian optimization, learning is performed when a new observation is obtained at each step.

\vspace{.1in}
\noindent {\bf Maximum likelihood estimation (MLE):} 
In Bayesian optimization, the objective function is modeled by a Gaussian process.
Therefore, given the outputs of the objective function obtained up to a certain step $\bm{y}=(y_1,\cdots,y_t)^T$, the log-likelihood of $\bm{y}$ can be calculated by:
\begin{align*}
    \log p(\bm{y}|\bm{\alpha}) \propto -\frac{1}{2}\log|\bm{K}+\sigma^2\bm{I}|-\frac{1}{2}\bm{y}^T(\bm{K}+\sigma^2\bm{I})^{-1}\bm{y}.
\end{align*}
We consider the use of maximum likelihood estimation to learn $\bm{\alpha}$, which maximizes this log-likelihood.
This can be performed by a gradient-based optimization method \cite{richard1995algorithms}. 
As in the case of kernel target alignment, we learn the coefficients when a new observation is obtained.

\section{Related work}
Bayesian optimization is widely used for optimizing expensive-to-evaluate, black-box, and noisy objective functions \cite{bo}.
For example, it is used for automated tuning of hyperparameters in machine learning models \cite{snoek2012bo}, path planning of mobile robots \cite{martinez2009bo} and finding the optimal set of sensors \cite{garnett2010bo}.
Although most studies on Bayesian optimization including these studies consider vectorial data, there are few studies that consider structured data such as graphs.

\par

The graph Bayesian optimization (GBO) was proposed as a framework of Bayesian optimization for graph data in particular for tree structred data  \cite{ramachandram2018bayesian}. Then, it was recently extended to an arbitrary graph structure \cite{cui2018graph}. 
GBO proposed by \cite{cui2018graph} uses a linear combination of two kernels.
One is a conventional vectorial kernel (e.g., linear kernel and Gaussian kernel) for the explicit feature vector including the number of nodes, average degree centrality, and average betweenness centrality.
The other one is a graph kernel, which may capture the implicit topological features that cannot be expressed by explicit features.
The coefficients of the linear combination is learned through the Bayesian optimization process.
After that, we can analyze which features expressed by the vectorial kernel or the graph kernel were effective as a result.
Specifically, they used the automatic relevance determination squared exponential (SEARD) kernel as a vectorial kernel and the deep graph kernel based on subgraphs \cite{deepgk} as a graph kernel.
However, to the best of our knowledge, there is no Bayesian optimization framework that \emph{explicitly} uses topological information.

\section{Experiments}
In this section, we evaluate our proposed algorithms using synthetic and two real datasets.

\subsection{Setup}
For the proposed method, we use maximum likelihood estimation like as described in Section \ref{mkl} for estimating the noise parameter $\sigma$ in Bayesian optimization.

\par
We set the hyperparameters of PWGK and PFK according to the original papers \cite{kusano2018kernel} and \cite{le2018persistence}, respectively.
Let $\set{D_1,\cdots,D_n}$ be the PDs for each point cloud in a dataset.
In PWGK, we use the weight function:
\begin{equation*}
	w(\boldx) = \arctan(C{\rm pers}(\boldx)^p),
\end{equation*}
where ${\rm pers}(\boldx)=d-b$ for $\boldx=(b,d)$. 
Therefore, the hyperparameters of PWGK-Linear are $C$ and $p$ in the weight function and the kernel bandwidth $\nu$. 
PWGK-Gaussian includes $\tau$ in addition. 
We fix the hyperparameters with the following values:
\begin{itemize}
	\item $C={\rm median}\set{{\rm pers}(D_i)|i=1,\cdots,n}$,
	\item $p=5$,
	\item $\nu={\rm median}\set{\nu(D_i)|i=1,\cdots,n}$,
	\item $\tau={\rm median}\Set{\left|\left|E(\mu_{D_i})-E(\mu_{D_j})\right|\right|_\calH|i<j}$,
\end{itemize}
where ${\rm pers}(D_i)={\rm median}\set{{\rm pers}(\boldx_j)|\boldx_j\in D_i}$ and $\nu(D_i)={\rm median}\set{\left|\left|\boldx_j-\boldx_k\right|\right| |\boldx_j,\boldx_k\in D_i,j<k}$.

\par
The hyperparameters of PFK are $\nu$ and $t$.
We search these parameters from $\nu\in\set{10^{-3},10,10^3}$ and $1/t\in\set{q_1,q_2,q_5,q_{10},q_{20},q_{50}}$, respectively, where $q_s$ is the s\% quantile of $\set{d_{FIM}(D_i,D_j)|i<j}$.

\par
We compare our proposed algorithm with the random search baseline and GBO \cite{cui2018graph}.
For GBO, since the synthetic data is given as a point cloud, we first compute a 5 nearest-neighbor graph and then feed the graph into GBO.
We use the same kernels as used in the original paper.
We extract 5 features from a graph (the number of nodes, the number of edges, average degree centrality, average betweenness centrality, and average clustering coefficient).
Each element $x$ is normalized by $\tilde{x}=(x-x_{min})/(x_{max}-x_{min})$.
The window size and embedding dimension for the deep graph kernel are chosen from $\set{2,5,10,25,50}$.
The kernel bandwidths in the SEARD kernel and the coefficients of the linear combination are estimated by maximum likelihood estimation.

In Bayesian optimization, we randomly choose 10 data points to calculate the predictive distribution for the first search point.
We use PWGK-Linear, PWGK-Gaussian and PFK as the kernel for PDs and EI as an acquisition function.
We first calculate the 1st PDs for synthetic dataset, and the 0th PDs for real datasets.
We calculate these kernels using approximation methods (random Fourier features for PWGK and fast Gauss transform for PFK, respectively).
We conduct Bayesian optimization 30 times. 

\begin{figure}[htbp]
	\centering
	\includegraphics[clip, width=0.48\textwidth, trim=2cm 10cm 2cm 11cm]{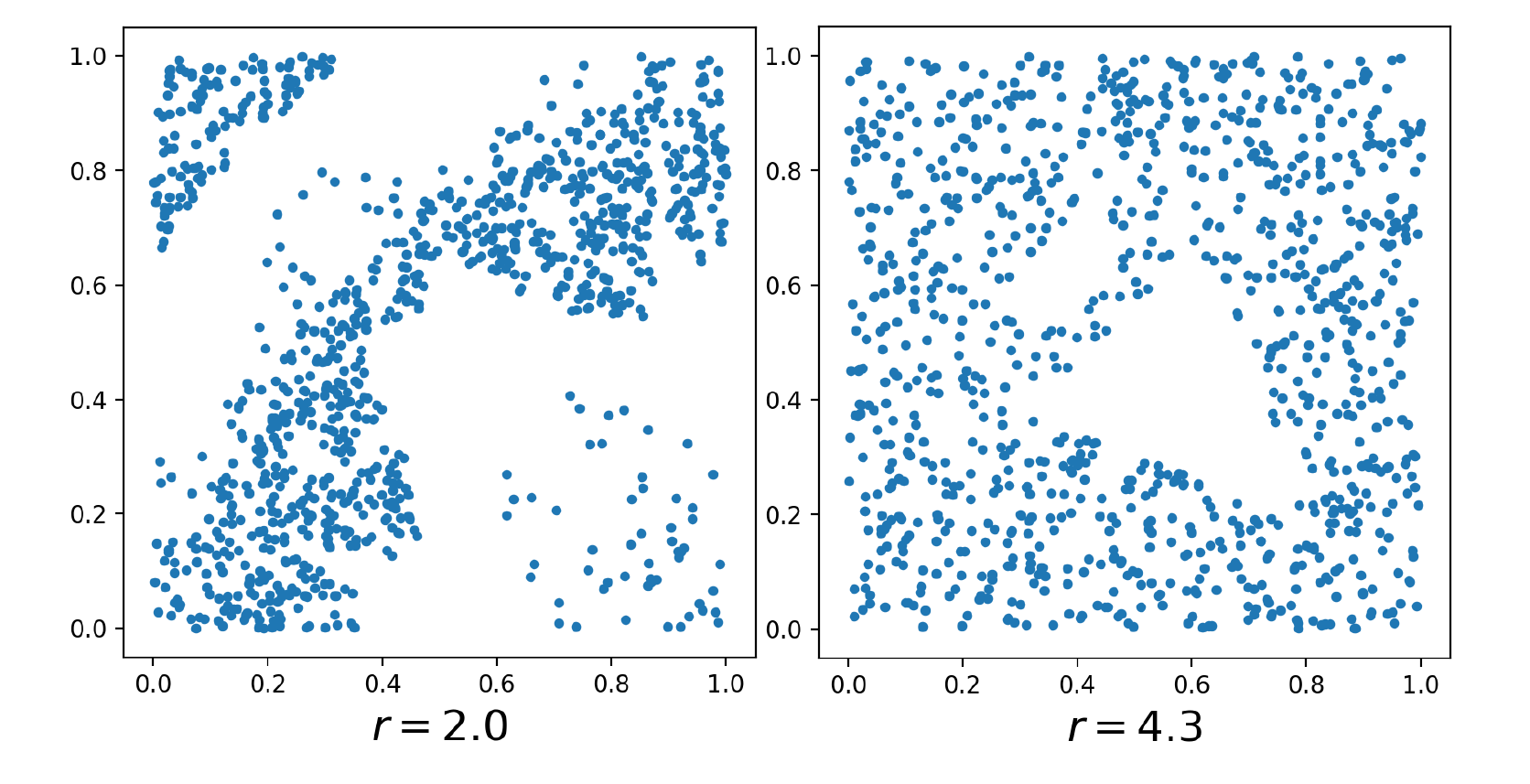}
	\caption{Illustrative examples of synthesized data.}
	\label{orbit_data}
	\vspace{-.2in}
\end{figure}

\begin{figure*}[htbp]
	\centering
	\includegraphics[clip, width=\textwidth, trim=2.75cm 13cm 2.75cm 13cm]{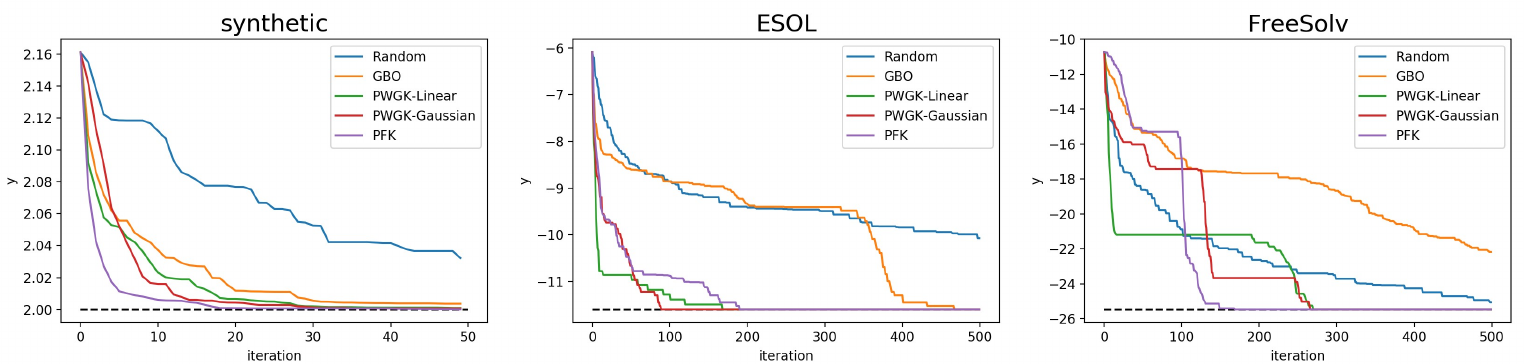}
	\caption{Comparison between random search baseline and PD kernels. The black dotted line shows the objective function value of the target data that we want to search for.}
	\label{exp1}
	\vspace{-.2in}
\end{figure*}

\subsection{Synthetic dataset}
To generate the synthetic dataset, we used the method proposed in \cite{orbit}.
This method generates a point cloud on $[0,1]\times[0,1]$.
We generate $M=1000$ point clouds consisting of $N=1000$ points as the dataset.
The specific procedure is as follows.
\begin{enumerate}
	\item Randomly choose $(x_0,y_0)\in[0,1]\times[0,1]$.
	\item Iterate the following procedure $M$ times.
	\begin{enumerate}
		\item Randomly choose $r\in[2.0,4.3]$.
		\item Generate a point cloud $\set{(x_1,y_1),\cdots,(x_N,y_N)}$ according to the following recurrence relations:
		\begin{eqnarray*}
			&& x_{n+1} = x_n+ry_n(1-y_n)\mod{1}, \\
			&& y_{n+1} = y_n+rx_{n+1}(1-x_{n+1})\mod{1}.
		\end{eqnarray*}
	\end{enumerate}
\end{enumerate}
The point clouds generated for $r=2.0$ and $r=4.3$ are shown in Figure \ref{orbit_data}.
We use the value of $r$, which was used to generate a point cloud, as the label of the point cloud.
In this study, we find the point cloud with minimum $r$ by using Bayesian optimization algorithms.

Figure \ref{exp1}-(synthetic) shows averages of the minimum observation obtained at each step for the synthetic data.
As we expected, the topological Bayesian optimization methods outperformed random search and the GBO algorithm. 

\subsection{Real datasets}
We used two real datasets about the properties of relatively small compounds from MoleculeNet \cite{moleculenet}.
ESOL is a dataset about the water solubility of 1128 compounds.
The average number of atoms is 25.6.
FreeSolv is a dataset about the hydration free energy of 642 compounds in water. 
The average number of atoms is 18.1.
For our method, we treat a compound as a point cloud using only the 3D coordinates of each atom forming the compound without considering any other information about atoms or bonds. 
We find the compound with minimum water solubility and hydration free energy by using the Bayesian optimization algorithms from the ESOL and FreeSolv datasets, respectively.

Figure \ref{exp1}-(ESOL)(FreeSolv) show averages of the minimum observation obtained at each step for real datasets.
In both cases, the information of PDs contributes to efficient search for the optimal structure.
Our method outperforms better in the case of the ESOL dataset than the case of the FreeSolv dataset. 
It may shows molecular structure reflects factors of water solubility than those of hydration free energy.



\subsection{Effectiveness of multiple kernel learning}
We compare Bayesian optimization using only one type of PD and that using combined multiple types of PD.
Here, we consider combining the information of the 0th PD and the 1st PD (i.e., $k=2$ in Eq.~\eqref{combination}).
We compare the cases of using only the 0th PD, using only the 1st PD and combining both information using kernel target alignment (align) and maximum likelihood estimation (MLE) as methods for learning the coefficients.
When combining the PFKs, we first conduct experiments similar to those in the previous section using only one type of PD and optimize the hyperparameters in PFKs, and in this experiment, we only learn the coefficients of a linear combination.

\par
The results are summarized in Table \ref{exp2}.
We evaluate the performances according to the area under the convergence curve.
That is, we calculate the area between the convergence curve and the black dotted line as shown in Figure \ref{exp1}.
The values in the table are scaled so that the case of random search baseline becomes 1.
In many cases, it is shown that the performance is improved by combining the information of both PDs by maximum likelihood estimation. 
In addition, when we apply PWGK-Linear to the ESOL dataset, the performance is better when combining by maximum likelihood estimation than when using only the 1st PD.
If there is no prior knowledge about which type of PD is effective, this shows that it may be better to combine both PDs than to choose one type of PD for intuition.
The same is true of applying PFK to the FreeSolv dataset.

\begin{table}[htbp]
\centering
\caption{Comparison between cases of using only one type of PD and of using multiple kernel learning methods.}
\begin{tabular}{|c|c|r|r|r|} \hline
    \multicolumn{2}{|c|}{} & Synthetic & ESOL & FreeSolv \\ \hline
    \multicolumn{2}{|c|}{Random} & 1.0000 & 1.0000 & 1.0000 \\ \hline
    \multicolumn{2}{|c|}{GBO} & 0.2157 & 0.6147 & 2.4099 \\ \hline
    PWGK & 0th & 0.1597 & {\bf 0.0571} & 0.6832 \\
    -Linear & 1st & 0.1551 & 0.3867 & 1.4169 \\
    & align & 0.1664 & 0.3119 & 1.0350 \\ 
    & MLE & {\bf 0.0898} & 0.1757 & {\bf 0.5241} \\ \hline
    PWGK & 0th & 0.1512 & 0.0763 & 0.8833 \\
    -Gaussian & 1st & {\bf 0.1509} & 0.4630 & 1.2399 \\
    & align & 0.1618 & 0.2455 & 0.8862 \\ 
    & MLE & 0.4308 & {\bf 0.0560} & {\bf 0.5867} \\\hline
    PFK & 0th & 0.1172 & 0.1153 & 0.7685 \\
    & 1st & {\bf 0.0730} & 0.2544 & {\bf 0.6644} \\
    & align & 0.0922 & 0.1195 & 0.8695 \\ 
    & MLE & 0.2220 & {\bf 0.0703} & 0.7640 \\\hline
\end{tabular}
\label{exp2}
\vspace{-.2in}
\end{table}

\section{Conclusion}
In this paper, we proposed the topological Bayesian optimization, which is a Bayesian optimization method using features extracted by persistent homology. 
In addition, we proposed a method to combine the kernels computed from multiple types of PDs by a linear combination, so that we can use the multiple topological features extracted from one source of data.
Through experiments, we confirmed that our method can search for the optimal structure from complex structured data more efficiently than the random search baseline and the state-of-the-art graph Bayesian optimization algorithm by combining multiple kernels using maximum likelihood estimation.


\bibliographystyle{named}
\bibliography{ijcai19}

\end{document}